\documentclass[runningheads]{llncs}
\usepackage{graphicx}
\usepackage{algorithmic}
\usepackage{float,xcolor}
\usepackage{algorithm}
\begin{document}
\title{VBSF-TLD: Validation-Based Approach for Soft Computing-Inspired Transfer Learning in Drone Detection}
%
%
\author{Jaskaran Singh\inst{1}\orcidID{0000-0001-9285-2048} }

\authorrunning{Singh et al.}
%
\institute{Department of Computer Science and Engineering\\Graphic Era Deemed to be University, Dehradun India 248002\\ \email{jaskaran.jsk2001@gmail.com }}

\maketitle              
\begin{abstract}

With the increasing utilization of Internet of Things (IoT)-enabled drones in diverse applications like photography, delivery, and surveillance, concerns regarding privacy and security have become more prominent. Drones have the ability to capture sensitive information, compromise privacy, and pose security risks. As a result, the demand for advanced technology to automate drone detection has become crucial. This paper presents a project on a transfer-based drone detection scheme, which forms an integral part of a computer vision-based module and leverages transfer learning to enhance performance. By harnessing the knowledge of pre-trained models from a related domain, transfer learning enables improved results even with limited training data. To evaluate the scheme's performance, we conducted tests on benchmark datasets, including the Drone-vs-Bird Dataset and the UAVDT dataset. Notably, the scheme's effectiveness is highlighted by its IOU-based validation results, demonstrating the potential of deep learning-based technology in automating drone detection in critical areas such as airports, military bases, and other high-security zones.

\end{abstract}
\section{Introduction}

The advancement of advanced drone technology has brought about a significant revolution in the field of aerial technology, making it more accessible and affordable for a wider range of consumers \cite{wulfovich2018drones}, \cite{floreano2015science}. However, along with these advancements, certain challenges, particularly pertaining to privacy concerns, have arisen. A typical drone comprises several key components, including the frame, propellers, brushless motors, flight control board, RF transmitter and receiver modules, camera unit, and rechargeable battery \cite{wz6}. In order to facilitate communication between drones and users via the Internet, an architecture known as the Internet of Drones (IoD), or IoT-enabled drones, has been developed \cite{wz5}. The availability of low-cost IoT-enabled drones equipped with advanced sensors and cameras has made it easier for individuals to capture and transmit images and videos without the knowledge or consent of others \cite{cavoukian2012privacy}. Consequently, concerns have been raised regarding the potential misuse of drones, particularly in sensitive areas such as law enforcement, where privacy holds paramount importance.

Machine Learning is a paradigm that involves the creation of algorithms and models allowing computers to learn from data and make predictions or decisions without explicit programming. By utilizing training and learning from past data, computers can enhance their performance. This field encompasses the development of mathematical models and algorithms capable of analyzing intricate data patterns, extracting valuable insights, and making predictions or taking actions based on these patterns. Machine Learning heavily relies on statistical techniques and computational algorithms to enable computers to continually learn and improve from data \cite{ml1}\cite{ml2}.

Computer vision is a subset of machine learning that deals with training machines using advanced algorithms and techniques to find patterns in visual images and videos. It uses visual clues to extract meaningful information from images and make decisions, detect objects, perform feature extraction, and interpretation. Computer vision is widely used in various industries and domains, including medical imaging for disease identification and treatment, autonomous vehicles, automated object detection modules, surveillance systems, augmented reality, and facial recognition systems \cite{cv1}\cite{cv2}\cite{cv3}. It is a rapidly emerging field that continues to evolve with the development of new pre-trained models, deep learning techniques, and the availability of large-scale visual datasets.

\subsection{Motivation of this work}
In recent years, the utilization of drones in illegal settings has significantly increased,
exploiting the lack of stringent regulations and unrestricted usage. This presents a formidable challenge for multinational corporations and private owners who must address the detection of these unauthorized drone activities as a matter of utmost importance. Currently, the prevailing approach involves hiring private security personnel, but this manual process is both costly and often ineffective.
\\ 
To tackle this challenge, there is a pressing need to automate the detection process through the utilization of AI algorithms and advanced camera systems. However, a significant gap exists in implementing deep learning software onto existing cameras and empowering them with the capability to accurately track and differentiate drones from other aerial objects in real-time. Bridging this gap would yield more efficient detection methods and facilitate the development of a detection-based alert system that seamlessly integrates into a centralized security framework.

\subsection{Research Contributions}

The research contributions of this paper are given below:
\begin{enumerate}
    \item In this paper we have proposed a validation based, soft computing inspired drone detection scheme VBSF-TLD.
    \item A generalised algorithm for the proposed VBSF-TLD scheme has been envisioned and formulated. 
    \item The practical effectiveness of the proposed VBSF-TLD scheme has been conducted on a curated dataset collected from benchmark resources. 
    \item The loss, accuracy, ROC curves and performance metrics of the VBSF-TLD scheme has been calculated to measure its detection capability.
\end{enumerate}

\section{Related Work}

In the present era, extensive research is being conducted to explore the use of computer vision libraries in the detection of moving objects during live feed deployment stages. This chapter reviews some of the significant existing work in this field, including state-of-the-art papers and datasets used for evaluating performance metrics. While a considerable amount of work has been done on the detection of drones using non-visual extraction methods such as radars and artificial neural networks \cite{i1}\cite{i2}\cite{i3}, relatively less emphasis has been placed on visually identifying drones in images using machine learning techniques with learned features. Firstly, all the papers reviewed in this chapter have either employed their own extracted data or utilized one of the three publicly available datasets. One of these datasets is the Drone vs Bird Dataset \cite{i4}, which presents a challenge in differentiating drones from birds. It consists of five videos comprising a total of 2727 frames with a resolution of 1920 × 1080 pixels. Another dataset, the UAV Project UAV dataset \cite{i5}, aims to address the scarcity of comprehensive UAV datasets. It is derived from 10 hours of raw videos, encompassing 80000 frames annotated with bounding boxes. This dataset stands out for its meticulousness and comprehensiveness, as it does not contain any synthetic data. Lastly, to differentiate drones from airplanes, several papers have utilized an airplane dataset \cite{i6}. This dataset consists of 10,200 images featuring 102 different variants of aircraft. Each image is labeled and annotated, making it a valuable resource for researchers working on drone detection.

Samadzadegan et al. \cite{14} propose a deep learning approach for efficient drone detection and recognition. They utilize the CSPDarknet53 feature extraction network and employ the Intersection over Union (IoU) loss function to differentiate between drones and birds. Their method successfully detects and distinguishes between two types of drones, while accurately differentiating them from birds. Al-Qubaydhi et al. \cite{15} implement an optimized version of YOLO (YOLOv5) to detect drones in videos with varying contrasts, including low contrast scenarios. Their dataset comprises drone images with diverse backgrounds, such as water, buildings, trees, and humans.

Wang et al. \cite{17} present a fast and cost-effective drone detection system using video images from static cameras. They employ temporal median background subtraction to identify moving objects in the video and extract global Fourier descriptors and local HOG features. These features are then used in an SVM classifier for classification and recognition. Xun et al. \cite{18} develop a drone surveillance system using the YOLOv3 model with pre-trained weights. The model is trained on a custom dataset and validated in real-time using an NVIDIA Jetson TX2 computing device.

Shi et al. \cite{19} propose a drone detection framework that utilizes three models - YOLOv4, YOLOv3, and SSD (Single-Shot-Detector) - with a CSPDarknet53 backbone network structure for real-time drone detection. These models are trained and tested on an augmented dataset that includes internet-collected images as well as their own collected images. A state-of-the-art paper \cite{20} introduces the use of a CNN model combined with a Harr feature cascade classifier for drone detection. The research sets a benchmark in this subdomain by training the network covertly with separate drone identification and detection modules. The achieved identification accuracy is 91.6\%, while the detection accuracy reaches 89\% over a manually gathered dataset. The Adam optimizer is employed in training the network.

\section{Proposed Methodology}

\subsection{Proposed system process flow}
The proposed workflow, as illustrated in Algorithm 1 and the architecture given in Figure\ref{figcond}, outlines the steps involved in creating an alarm detection system for identifying drones in the airspace. The system utilizes various components, including a CCTV camera connected to a computing server, an AlexNet Background Subtractor, a GAN-based Super Resolution Module, and a Transfer Learning Drone Detection Model (FastRCNN ). The desired output is a reliable drone detection system that triggers an alert when drones are detected.

\begin{figure}[H]
\centerline{\includegraphics[width=87mm,scale=1]{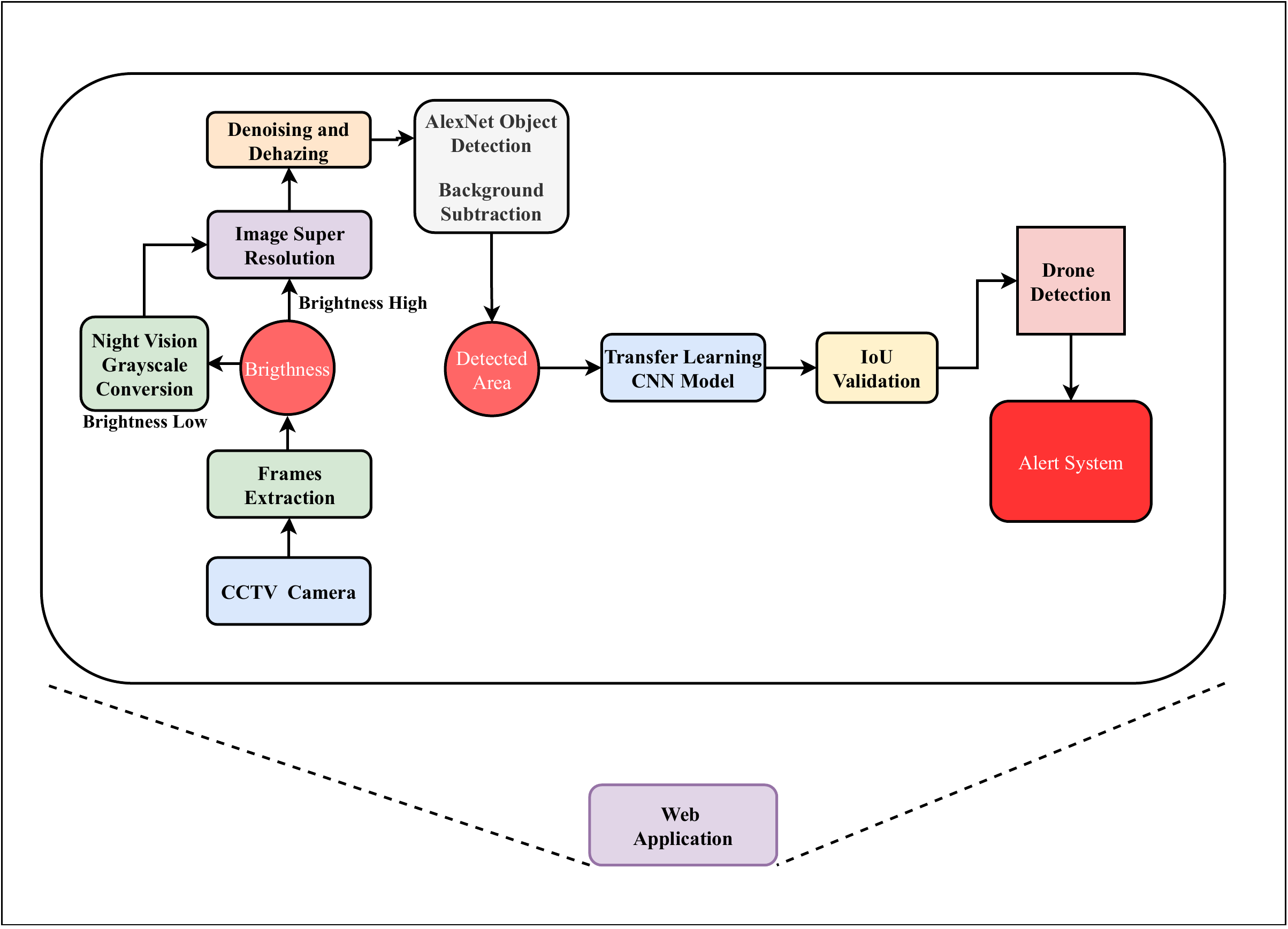}}
\caption{Process Flow of Proposed VBSF-TLD Scheme.}
\label{figcond}
\end{figure}

\begin{algorithm}
  \caption{Drone Detection System}
  \textbf{Input:} A CCTV  camera that is connected to a compting server,\\
An AlexNet Background Subtractor, \\
A GAN Based Super Resolution Module, \\
 A Transfer Learning Drone Detection Model (FastRCNN)\\
\textbf{Output:} An alarm detection system, that detects drones in the air space\\
  \label{algorithm}
  \begin{algorithmic}[1]
    \STATE Set the system to run for a total duration of 30 minutes.
    \FOR{$i=1$ to $10$}
      \STATE Set the current time as the starting time.
      \WHILE{current time is less than 10 minutes after the starting time}
        \STATE Extract image frames from the installed CCTV camera.
        \STATE Calculate the brightness of the extracted image using the alpha channel.
        \IF{brightness is below the threshold}
          \STATE Apply night vision grayscale conversion to the image.
          \STATE Apply image super-resolution using a GAN-based enhancing technique.
          \STATE Perform denoising and dehazing on the image to increase object clarity.
          \STATE Apply pretrained UNET object detection and background subtraction module to remove the background and retain the foreground objects.
          \STATE Use the transfer learning-based CNN model (FastRCNN) to detect drones.
          \STATE Validate the results by performing Intersection over Union (IoU) validation with the subsequent frame.
          \IF{detection is consistent}
            \STATE Trigger the IoT-based Alert System for drone detection.
          \ELSE
            \STATE Move to the next frame.
          \ENDIF
        \ELSE
          \STATE Skip to the next frame.
        \ENDIF
        \STATE Update the current time.
      \ENDWHILE
      \STATE Wait for 30 minutes before starting the loop again.
    \ENDFOR
  \end{algorithmic}
\end{algorithm}

\subsection{Data Collection and Augmentation}

For the data acquisition part, we collected samples of aerial objects from available datasets, as well as custom-collected videos from various sources such as documentaries, vlogs, and movies. We collected two broad types of videos. The first type consists of videos in which drones and quadcopters are being flown, with the camera remaining static to mimic the type of camera that will be present in our project. The second type includes various videos of aerial objects like birds and airplanes, which will help the model better differentiate drones from other aerial objects. Our dataset includes references from papers \cite{i4}\cite{i5}\cite{i6}. 	

Using frame extraction , we obtained 650 frames of drone footage and 800 frames of normal aerial objects from the existing dataset in a usable format. In addition to this, we curated our dataset by employing web scraping techniques and searching for images in blogs, reviews, Kaggle datasets, and manually collecting videos and images from various sources, including documentary blogs, movies, and commercials, over a period of one month. We then carefully screened and performed a manual annotation process on this new data, resulting in the creation of our proprietary dataset. This dataset consists of over 2800 samples, including 1350 samples of drones and quadcopters, and 1450 samples of real objects. Our dataset is highly diverse, including drones and quadcopters from multiple brands, homemade variations, and commercial drone delivery vehicles, as well as airplanes, birds, helicopters, and balloons as representative objects.

Finally, we performed thorough data augmentation to enhance training and generate new data. This involved modifying the images by removing noise, haze, rain, and fog. We also adjusted the brightness and varied the light conditions. Furthermore, we applied filtering techniques to enhance or remove specific color bands. Custom cropping of the images was performed to select specific regions, and flipping, rotating, and scaling techniques were applied to account for drone positioning and create variations in the dataset.

\begin{figure}[H]
\centerline{\includegraphics[width=87mm,scale=1]{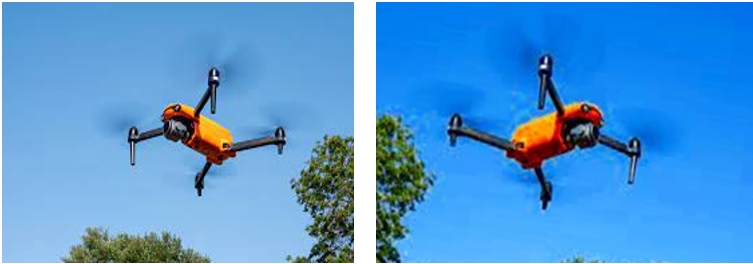}}
\caption{Results of Data Augmentation: (Left) Original Image, (Right) Augmented Image}
\label{figcond}
\end{figure}

\subsection{ESRGAN-based Super Resolution Module}

n order to further enhance the quality of the training dataset, we utilized an Enhanced Super-Resolution Generative Adversarial Network (ESRGAN). This advanced technique improved the dataset by augmenting the images and effectively eliminating any existing noise, resulting in remarkable improvements in clarity and overall quality. ESRGAN builds upon the ResNet-style architecture of SRGAN, but with a notable enhancement: it replaces the residual block with the Residual-in-Residual Dense Block. Through an adversarial training process, ESRGAN incorporates both a generator network and a discriminator network. The generator network's objective is to generate high-resolution representations of low-resolution input images that closely resemble the original low-resolution images. Simultaneously, the discriminator network is trained to distinguish between artificially generated high-resolution images and genuine high-resolution photographs. As the training progresses iteratively, the discriminator network becomes increasingly adept at recognizing the generated images, while the generator network learns to produce higher quality and more realistic high-resolution images. Our findings indicate that ESRGAN surpasses Super Resolution GAN (SRGAN) in terms of both accuracy and time complexity, establishing it as a significantly superior approach.

\subsection{Background Subtraction Module}

To perform background subtraction of drone images a pre-trained U-Net \cite{24} model was used. Initially, the U-Net model was loaded and pre-trained weights were downloaded, leveraging the transfer learning paradigm to reduce training time. The U-Net architecture consists of an encoder and a decoder. The encoder gradually downsampled the input image to capture image-level features, while the decoder gradually upsampled the features to generate a segmentation mask. This segmentation mask could then be overlapped with the original image to remove the background.
Subsequently, a dataset of drone images was collected and labelled as either foreground or background, where the background represented areas of the image without the drone. The U-Net model was fine-tuned on this labelled dataset using a Stochastic Gradient Descent optimizer and a binary cross-entropy loss function, for the updating of model weights. Once the model was trained, it was employed to perform background subtraction on new drone images. Each image was passed through the model, and the foreground pixels were extracted. The resulting foreground pixels represented the drone, while the remaining background pixels corresponded to the scene behind the drone, as indicated by the generated segmentation mask. This segmentation assists in the effective tracking and monitoring of the drone's movements.

\subsection{FastRCNN-based Object Detection Model}

We utilized the FastRCNN \cite{25} model, a highly effective approach for object detection, in the development of our object detection model. The FastRCNN model employs selective search to generate object proposals. These proposals are then fed into a pre-trained convolutional neural network, specifically VGG19, to extract features. The extracted features are used for classifying the proposals into either a drone or a non-drone object.
Here, we froze all layers except the last two. This freezing prevents these layers from being updated during the training process, allowing the model to retain the knowledge captured by the pre-trained convolutional neural network. Finally, we added a dense layer for binary sigmoid-based classification. This additional layer enables the model to assign a probability score to each input, indicating the likelihood of it being a drone or a non-drone object. The sigmoid activation function ensures that the output falls within the range of 0 to 1, representing the probability of being a drone.
To optimize the performance of our FastRCNN model for drone detection, we employed various techniques. Firstly, we fine-tuned the model on our drone dataset by adjusting hyperparameters, including learning rate and regularization parameters. To mitigate the risk of overfitting, which occurs when the model performs well on training data but poorly on new data, we incorporated additional dropout and batch normalization layers. Dropout randomly drops out some neurons during training, acting as a form of regularization. Batch normalization normalizes the input to each layer, promoting stable gradient flow and faster training. We experimented with different dropout rates and batch normalization parameters to identify the optimal combination that reduced overfitting and improved the model's validation accuracy.

\subsection{Intersection over Union Validation Module}

To validate our results, we implemented a dedicated section aimed at assessing the accuracy of the bounding box predictions generated by our tracking algorithm. This section served as an external module, separate from the training process, focusing solely on evaluating the performance of our algorithm. To measure accuracy, we used the concept of the area of intersection, which calculates the overlap between the predicted bounding box and the ground truth bounding box. By calculating the ratio of the intersection area to the union area, known as the Intersection over Union (IoU), we obtained a measure of accuracy.

To determine a threshold for successful detection, we set it at 90\%. If the IoU value for a specific frame exceeded this threshold, it indicated a highly accurate prediction with a significant overlap between the predicted and ground truth bounding boxes. On the other hand, if the IoU value fell below the threshold, it suggested a less accurate prediction with limited overlap or a potential false positive. We chose the 90\% threshold based on our application's specific requirements, aiming to balance between false positives and false negatives.

To evaluate the algorithm's performance, we analyzed the percentage of frames where the predicted bounding boxes met or exceeded the 90\% IoU threshold. This allowed us to determine the overall accuracy and reliability of the tracking algorithm. Additionally, we visually inspected the overlap between the predicted and ground truth bounding boxes in these frames to verify the correctness of the detections. This visual inspection provided further assurance and qualitative assessment of the algorithm's ability to accurately track the drone's bounding box.

\subsection{Soft Computing-Inspired Particle Swarm Optimization }

The research paper presents a novel approach for drone detection, leveraging a soft computing-inspired optimization function based on Particle Swarm Optimization (PSO) \cite{pso}. Traditional approaches often struggle with the uncertainties and complexities inherent in drone detection tasks. To address this, the proposed method incorporates PSO, a nature-inspired optimization algorithm that mimics the collective behavior of a swarm of particles. By integrating the PSO optimizer into the FastRCNN model, the loss function is dynamically optimized, allowing the model to adaptively adjust its parameters during training. The soft computing-inspired loss function considers the localization accuracy, object confidence, and class probabilities, aiming to enhance the model's performance in detecting drones accurately. The PSO algorithm works principally on  these equations:

\textbf{Updation of Particle Velocity:}
\begin{equation}
v_{ij}(t+1) = \omega \cdot v_{ij}(t) + c_1 \cdot r_1 \cdot (pbest_{ij}(t) - x_{ij}(t)) + c_2 \cdot r_2 \cdot (gbest_{j}(t) - x_{ij}(t))
\end{equation}

where \(v_{ij}(t+1)\) is the updated velocity of particle \(i\) in dimension \(j\) at time \(t+1\), \(\omega\) is the inertia weight, \(v_{ij}(t)\) is the current velocity of particle \(i\) in dimension \(j\) at time \(t\), \(c_1\) and \(c_2\) are the cognitive and social acceleration coefficients, respectively, \(r_1\) and \(r_2\) are random numbers between 0 and 1, \(pbest_{ij}(t)\) is the best position found by particle \(i\) in dimension \(j\) up to time \(t\), and \(gbest_{j}(t)\) is the best position found globally in dimension \(j\) up to time \(t\).

\textbf{Updation of Particle Position:}
\begin{equation}
x_{ij}(t+1) = x_{ij}(t) + v_{ij}(t+1)
\end{equation}

where \(x_{ij}(t+1)\) is the updated position of particle \(i\) in dimension \(j\) at time \(t+1\), and \(v_{ij}(t+1)\) is the updated velocity of particle \(i\) in dimension \(j\) at time \(t+1\).

These equations are the main update rules of Particle Swarm Optimization, where each particle's velocity is updated based on its previous velocity, the best position it has achieved (pbest), and the best position achieved globally (gbest). The updated velocity is then used to update the particle's position, allowing it to effeciently explore the search space.

\section{Practical Implementation}
This section provides practical implementation details for the proposed VBSF-TLD, including information on the hardware and software employed. Our hardware setup consisted of a processor featuring 2 X Intel(R) Xeon(R) with a clock speed of 2.20GHz, along with 12 GB of RAM and a Nvidia Tesla T4 GPU. The training was conducted on the Google Colab platform, utilizing the Ubuntu 18.04.5 LTS platform for simulation purposes.

For the implementation of our scheme, we utilized Python 3.8 as the programming language, along with the Tensorflow library with Keras API. The model was trained and validated on the collected dataset, which was partitioned accordingly. We employed K4 cross-validation, dividing the dataset into four equal folds. Each fold was used to train and test the model, resulting in an average accuracy assessment. To compile the model, we utilized the PSO (Particle Swarm Optimization) optimizer and binary cross-entropy loss function, training the model for 50 epochs. Throughout the training process, we closely monitored various performance metrics, including accuracy, loss, and validation accuracy, ensuring that the model was learning effectively without overfitting to the training data.

Table \ref{mlresults2} presents a summary of the performance metrics attained by the proposed VBSF-TLD, including accuracy, recall, precision, and F1-score.

\begin{table}[!htb]
\caption{Performance of the proposed VBSF-TLD Scheme}
\label{mlresults2}
\centering
\begin{tabular}{|p{1.8cm}|p{1.2cm}|p{1.2cm}|p{1.5cm}|}
\hline
\textbf{Accuracy} & \textbf{Precision} & \textbf{Recall} & \textbf{F1-Score} \\ \hline
\ 91.36\%      &      89.39\%        &           92.58\%      &      90.96\%   \\ \hline
\end{tabular}
\end{table}

\begin{figure*}[!htb]
\centering
\includegraphics[width=0.59\linewidth]{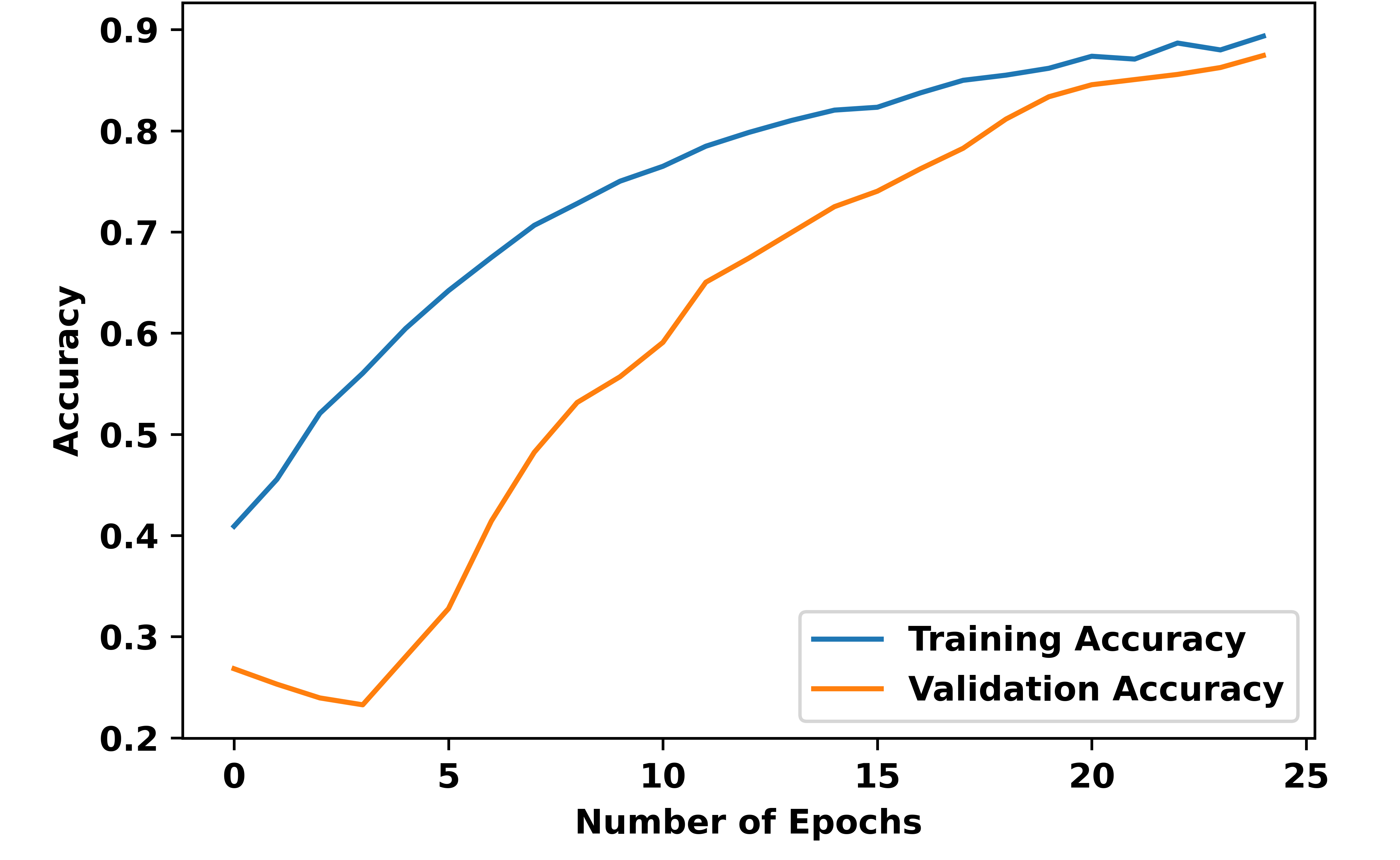}
\caption{Accuracy curve of the proposed VBSF-TLD}  \label{acc}
\end{figure*}

\begin{figure*}[!htb]
\centering
\includegraphics[width=0.59\linewidth]{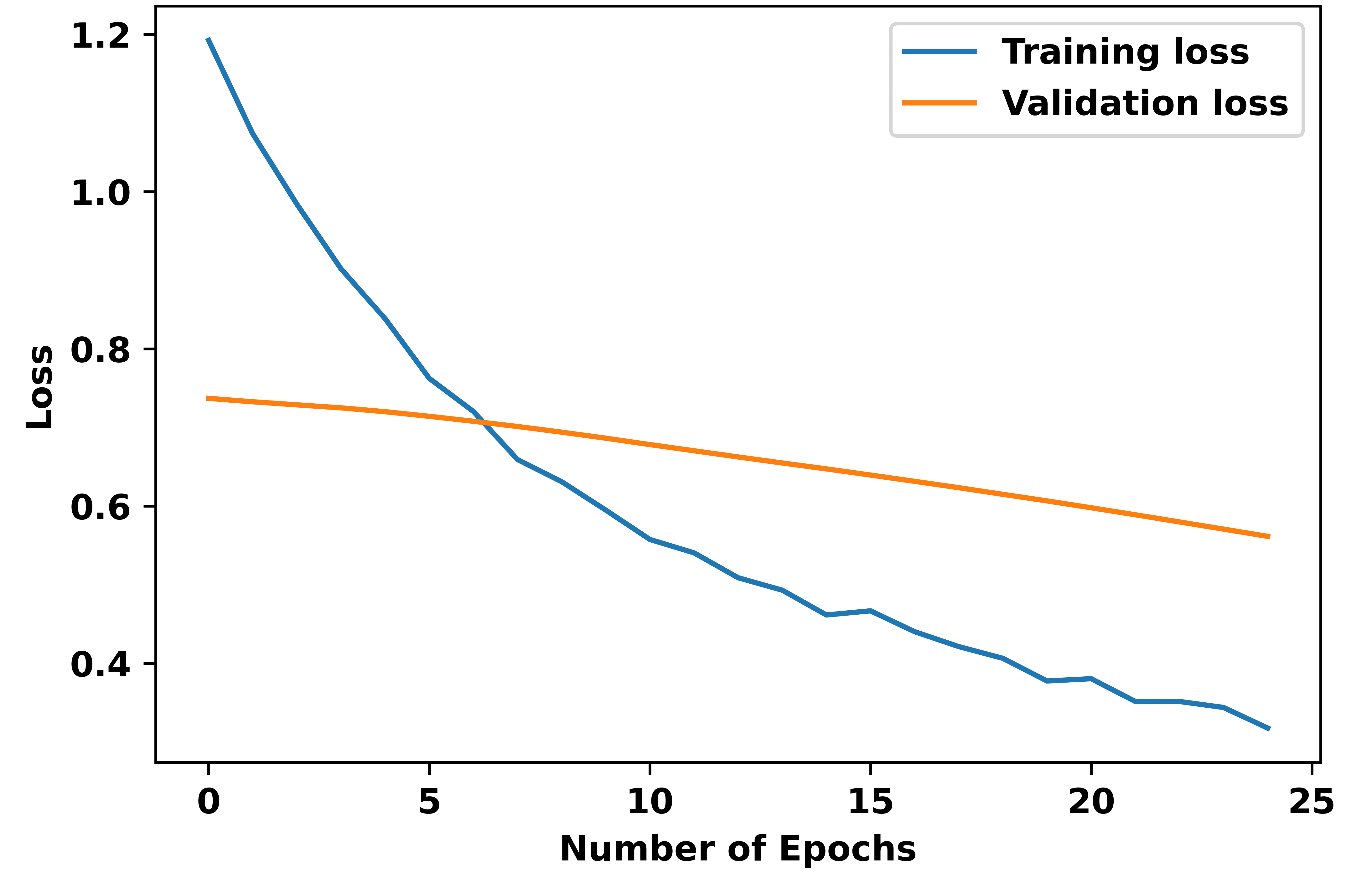}
\caption{Loss curve of the proposed VBSF-TLD}  \label{loss}
\end{figure*}

Figure \ref{acc}  illustrates the accuracy, while Figure\ref{loss}  visualizes the loss during the training and validation process, facilitating a clear comprehension of the VBSF-TLD drone detection scheme.

\section{Conclusion}

In this paper, we have presented a novel approach for detecting drones using transfer learning and background subtraction techniques, which is automated and cost-effective. Our results demonstrate the effectiveness of this approach in achieving accurate and precise drone detection. The proposed model incorporates rigorous processing techniques, including a GAN-based approach and background subtraction, eliminating the need for specialized hardware. This capability allows us to enhance image quality, removing moving noise caused by varying atmospheric conditions in practical scenarios, without requiring substantial investments in cameras and filters. By leveraging these advantages, we believe that our proposed Validation-Based Approach for Soft Computing-Inspired Transfer Learning in Drone Detection Scheme (VBSF-TLD) holds significant potential for enhancing security and safety against unauthorized drone flights in various day-to-day applications.

In future, we would like to extend proposed VBSF-TLD by incorporating advanced machine learning techniques such as ensemble deep learning and reinforcement learning. This will further enhance the accuracy and efficiency of our approach, enabling even more robust drone detection in diverse environments.

%
%
 \bibliographystyle{splncs04}
 \bibliography{ref}

\begin{thebibliography}{00}




\bibitem{wulfovich2018drones} Wulfovich, S., Rivas, H., and Matabuena, P. Drones in healthcare. Digital Health: Scaling Healthcare to the World, 159-168, 2018.
\bibitem{floreano2015science} Floreano, Dario, and Robert J. Wood. "Science, technology and the future of small autonomous drones." nature 521.7553, 2015.
\bibitem{wz6} Boccadoro, P., Striccoli, D. and Grieco, L.A. An extensive survey on the Internet of Drones. Ad Hoc Networks, 122, 2021.
\bibitem{wz5} Internet of drones security: Derhab, A., Cheikhrouhou, O., Allouch, A., Koubaa, A., Qureshi, B., Ferrag, M.A., Maglaras, L. and Khan, F.A. Internet of Drones Security: Taxonomies, Open Issues, and Future Directions. Vehicular Communications, p.100552, 2022.
\bibitem{cavoukian2012privacy} Cavoukian, Ann. Privacy and drones: Unmanned aerial vehicles. Ontario: Information and Privacy Commissioner of Ontario, Canada, 2012.
\bibitem{ml1} Bishop, Christopher M., and Nasser M. Nasrabadi. Pattern recognition and machine learning. Vol. 4. No. 4. New York: Springer, 2006.
\bibitem{ml2} Jordan, Michael I., and Tom M. Mitchell. "Machine learning: Trends, perspectives, and prospects." Science 349.6245, 2015.
\bibitem{cv1} Gao, J., Yang, Y., Lin, P., and Park, D. S. Computer vision in healthcare applications. Journal of healthcare engineering, 2018.
\bibitem{cv2} Cazzato D, Cimarelli C, Sanchez-Lopez JL, Voos H, Leo M. A Survey of Computer Vision Methods for 2D Object Detection from Unmanned Aerial Vehicles. Journal of Imaging, 2020.
\bibitem{cv3} Jindal, Vishesh, Shailendra Narayan Singh, and Soumya Suvra Khan. "Facial Recognition with Computer Vision." Machine Intelligence and Data Science Applications: Proceedings of MIDAS 2021. Singapore: Springer Nature Singapore, 2022. 313-330.

\bibitem{i1} N.M. Jahangir and C. Baker, "Robust detection of micro-UAS drones with L-band 3-D holographic radar," in Proc. IEEE Sensor Signal Process. Defence (SSPD), Sep. 2016, pp. 1–5.
\bibitem{i2} B. Torvik, K. E. Olsen, and H. Griffiths, "Classification of birds and UAVs based on radar polarimetry," IEEE Geosci. Remote Sens. Lett., vol. 13, no. 9, pp. 1305–1309, Sep. 2016.
\bibitem{i3} D. S. Zrnić and A. V. Ryzhkov, "Observations of insects and birds with a polarimetric radar," IEEE Trans. Geosci. Remote Sens., vol. 36, no. 2, pp. 661–668, Mar. 1998.
\bibitem{i4} A. Coluccia et al., "Drone-vs-Bird Detection Challenge at IEEE AVSS2021," 2021 17th IEEE International Conference on Advanced Video and Signal Based Surveillance (AVSS), 2021,pp. 1-8.
\bibitem{i5} D. Du, Y. Qi, H.g Yu, Y. Yang, K. Duan, G. Li, W.g Zhang, Q. Huang, Q. Tian, " The Unmanned Aerial Vehicle Benchmark: Object Detection and Tracking", European Conference on Computer Vision (ECCV), 2018.
\bibitem{i6} S. Maji, J. Kannala, E. Rahtu, M. Blaschko, A. Vedaldi, (2013). Fine-Grained Visual Classification of Aircraft [Techreport].



\bibitem{14} F. Samadzadegan, F. Dadrass Javan, F. Ashtari Mahini, M. Gholamshahi, Detection and recognition of drones based on a deep convolutional neural network using visible imagery, Aerospace 9, 31, 2022. 
\bibitem{15} N. Al-Qubaydhi, A. Alenezi, T. Alanazi, A. Senyor, N. Alanezi, B. Alotaibi, M. Alotaibi,   A. Razaque, A. A. Abdelhamid, A. Alotaibi, Detection of unauthorized unmanned aerial vehicles using yolov5 and transfer learning, Electronics 11, 2669, 2022.
\bibitem{17} Z. Wang, L. Qi, Y. Tie, Y. Ding, Y. Bai, Drone detection based on fd-hog descriptor, in: 2018 International Conference on Cyber-Enabled Distributed Computing and Knowledge Discovery (CyberC), 2018.
\bibitem{18}D. T. Wei Xun, Y. L. Lim, S. Srigrarom, Drone detection using yolov3 with transfer learning on nvidia jetson tx2, in: 2021 Second International Symposium on Instrumentation, Control, Artificial Intelligence, and Robotics (ICA-SYMP), 2021.
\bibitem{19} Q. Shi, J. Li, Objects detection of uav for anti-uav based on yolov4, in: 2020 IEEE 2nd International Conference on Civil Aviation Safety and Information Technology (ICCASIT, 2020.
\bibitem{20} D. Lee, W. G. La, and H. Kim, ‘‘Drone detection and identification system using artificial intelligence,’’ in Proc. Int. Conf. Inf. Commun. Technol. Converg. (ICTC), Oct. 2018.






\bibitem{24} O. Ronneberger, F. Philipp, and T. Brox. "U-net: Convolutional networks for biomedical image segmentation." Medical Image Computing and Computer-Assisted Intervention–MICCAI 2015: 18th International Conference, Munich, Germany, October 5-9, 2015, Proceedings, Part III 18. Springer International Publishing, 2015.
\bibitem{25} R. Girshick, "Fast R-CNN," 2015 IEEE International Conference on Computer Vision (ICCV), Santiago, Chile, pp. 1440-1448, doi: 10.1109/ICCV.2015.169, 2015.
\bibitem{pso} Kennedy, James, and Russell Eberhart. "Particle swarm optimization." Proceedings of ICNN'95-international conference on neural networks. Vol. 4. IEEE, 1995.

\end{thebibliography}

\end{document}